\documentclass{article} % For LaTeX2e
\usepackage{iclr2025_conference,times}

\usepackage{amsmath,amsfonts,bm}

\def\eqref#1{equation~\ref{#1}}
\def\1{\bm{1}}

\DeclareMathAlphabet{\mathsfit}{\encodingdefault}{\sfdefault}{m}{sl}
\SetMathAlphabet{\mathsfit}{bold}{\encodingdefault}{\sfdefault}{bx}{n}

\definecolor{RoyalBlue}{RGB}{40,155,230}
\usepackage{cases}
\usepackage{bm}
\usepackage{color}
\usepackage{graphicx}
\usepackage{subfigure}
\usepackage{wrapfig}
\usepackage{booktabs}
\usepackage{multirow}
\usepackage{multicol}
\usepackage{algorithm}
\usepackage{algorithmic} 
\usepackage{bbm}
\usepackage{url}
\usepackage{multirow}
\usepackage{multicol}
\usepackage{makecell}
\usepackage[utf8]{inputenc}   % 确保输入编码为 UTF-8
\usepackage[T1]{fontenc}      % 使用 T1 字体编码（支持更多符号）
\usepackage{textcomp}
\usepackage{shellesc} % 启用 shell 逃逸

\usepackage{tcolorbox}
\tcbset{
  casebox/.style={
    colback=gray!10,  % 背景色（浅灰）
    colframe=black, % 边框色（浅蓝）
    rounded corners,  % 圆角
    boxrule=0.5pt,    % 边框粗细
    left=10pt, right=10pt, top=5pt, bottom=5pt  % 内边距
  }
}

\usepackage[
  fencedCode=false
]{markdown}

\usepackage{xcolor}

\title{Zero Reinforcement Learning Towards General Domains}

\author{
Yuyuan Zeng$^1$, Yufei Huang$^{1}$, Can Xu$^{1}$, Qingfeng Sun$^{1}$, Jianfeng Yan$^{1}$, Guanghui Xu$^{1}$, Tao Yang$^{1}$, Fengzong Lian$^1$ \\
\vspace{2mm}
\textbf{$^1$LLM Department, Tencent}
}

\begin{document}

\maketitle

\begin{abstract}
Zero Reinforcement Learning (Zero-RL) has proven to be an effective approach for enhancing the reasoning capabilities of large language models (LLMs) by directly applying reinforcement learning with verifiable rewards on pretrained models, without the need for a supervised fine-tuning phase. However, current research on zero-RL primarily focuses on domains with easily verifiable reward signals, such as mathematics, programming, and other reasoning tasks. The challenge of eliciting reasoning abilities in more diverse scenarios, where verification is not straightforward, remains underexplored. To address this gap, we propose a novel zero-RL paradigm designed to improve a model's reasoning ability across both verifiable and non-verifiable domains. By combining verifiable rewards with a generative reward model, we conduct multi-task zero-RL training across both domains, facilitating the transfer of reasoning capabilities between them. Furthermore, to mitigate reward hacking in the generative reward model, we design a smooth length penalty that encourages the generation of more comprehensive thinking tokens in general domains. Experimental results on Qwen3-8B-Base and Qwen3-14B-Base demonstrate that our approach achieves superior reasoning performance, not only on tasks requiring extensive reasoning but also on more general tasks.
\end{abstract}
% Model-based reward designed to improve models’ capabilities in diverse scenarios suffers from reward hacking problems during RL training, which hinders large-scale reinforcement learning training.
% the effectiveness of our method on various benchmarks (e.g. MMLU-Pro, SuperGPQA, MATH-500, Arena-Hard etc.) covering wide general domains. The evaluation results demonstrate that our method can surpass the other RL-Zero counterparts achieving superior generalization ability.

\section{Introduction}
Recent advances in large language models (LLMs) have unlocked significant potential for artificial intelligence across diverse domains. In particular, efforts such as OpenAI-o1~\citep{OpenAI} and DeepSeek-R1~\citep{guo2025deepseek} have introduced long chain-of-thought (CoT) reasoning, which represents a breakthrough in enhancing the reasoning capabilities of LLMs. A central technique in this progress is zero reinforcement learning proposed in DeepSeek-R1-Zero~\citep{guo2025deepseek}, which converts large language models into large reasoning models (LRMs). Specifically, zero-RL directly trains pretrained LLMs using reinforcement learning algorithms such as Group Relative Policy Optimization (GRPO)~\citep{shao2024deepseekmath} and Proximal Policy Optimization (PPO)~\citep{ppo} with verifiable reward signals, and has demonstrated remarkable improvements on verifiable domains like mathematics and programming~\citep{yu2025dapo, zeng2025simplerl}.

There has been a surge of interest in expanding zero-RL methods to more diverse reasoning tasks, as exemplified by General-Reasoner~\citep{ma2025general} and Nemotron-Crossthink~\citep{akter2025nemotron}. However, existing work on general zero-RL-based reasoning still largely focuses on STEM domains, where problems have well-defined ground truths that allow straightforward verification of model outputs. In contrast, applying zero-RL to open-ended tasks remains underexplored. The key challenge lies in the fact that responses in open-ended tasks are difficult to verify, making it hard to obtain reliable and specific reward signals.

% However, zero reinforcement learning in general domain is rarely explored. Recent works in the field of zero-RL, such as General-Reasoner~\cite{ma2025general}, Nemotron-Crossthink~\cite{akter2025nemotron} are proposed to enhance the reasoning capabilities of LLMs across diverse reasoning domains (e.g., STEM tasks) beyond mathematics by model-based verifier and blending multi-domain corpora, leaving other open-ended tasks via zero-RL not well addressed. To improve the capabilities of LLMs across multiple domains is challenging, because the reward signals for open-ended tasks is hard to model. Model-based reward such as generative reward model (GRM) ~\cite{mahan2024generative} is generally exploited, however it often suffers from reward hacking problems which hinders large-scale reinforcement learning.

To address these challenges, we propose a unified zero-RL framework (General Zero-RL) which integrates both verifiable rewards and generative reward models. Unlike conventional zero-RL training on reasoning data, our preliminary experiments reveal that training solely on general-domain data struggles to elicit meaningful reasoning behaviors (e.g., step-by-step analysis, self-reflection, and backtracking). In such cases, the model often tends to output final answers without a substantive thinking process, or with only superficial reasoning steps. To mitigate this, we adopt a multi-task zero-RL training strategy that transfers reasoning behaviors across diverse domains by jointly training on both general and reasoning data.

Specifically, for reasoning tasks with ground-truth answers (treated as verifiable tasks), we use the final accuracy assessed by a verifier model as the reward signal. For general tasks with open-ended answers, we employ a generative reward model that evaluates the quality of responses and provides corresponding reward scores. However, since model-based reward models are often criticized for favoring longer responses~\citep{DBLP:journals/corr/abs-2310-10076, DBLP:journals/corr/abs-2407-01085}, we observe that on general data the model tends to produce unnecessarily lengthened answers rather than substantive reasoning processes. To mitigate this reward hacking issue, we introduce a smooth length penalty, which penalizes the length difference between the length of the reasoning process and the answer content to prevent the model from producing verbose response.
% inconsistency between the length of the reasoning process and summary. 
This length penalty regularization yields a more stable increase in response length during RL training. Furthermore, by gradually expanding the maximum allowable response length during training~\citep{wang2025octothinker}, we can avoid sudden spikes in response length and further stabilize model optimization.

To assess the effectiveness of our approach, we conduct extensive experiments on Qwen3-8B-Base and Qwen3-14B-Base~\citep{yang2025qwen3}, and evaluate our models across three categories of tasks: \textbf{Math Reasoning}, including MATH-500~\citep{lightman2023let}, AIME~\citep{AIME}, and OlympiadBench~\citep{he2024olympiadbench}; \textbf{General Reasoning}, including MMLU-Pro~\citep{wang2024mmlu}, GPQA-Diamond~\citep{rein2024gpqa}, SuperGPQA~\citep{du2025supergpqa} and BBEH~\citep{kazemi2025big}; and \textbf{General Tasks}, such as Arena-Hard~\citep{li2024crowdsourced}, WritingBench~\citep{wu2025writingbench}, WildBench v2~\citep{lin2024wildbench} and AlpacaEval2.0~\citep{dubois2024length}. On reasoning tasks, our General Zero-RL models not only outperform other zero-RL trained models of comparable size, but also surpass DeepSeek-R1-Zero-Qwen-32B~\citep{guo2025deepseek} by a significant margin. On general tasks, which have received little attention in prior zero-RL research, our models also achieve competitive results and generate coherent, meaningful reasoning content. Ablation studies further confirm the crucial role of our multi-task learning and length penalty strategy. Collectively, these findings demonstrate that our multi-task zero-RL training effectively enhances the reasoning capabilities of pretrained LLMs across both verifiable and non-verifiable tasks.

% The performance of our General-Zero-Qwen3-8 surpasses General-Reasoner-7B~\citep{ma2025general}, Nemotron-CrossThink-7B~\citep{akter2025nemotron} and comparable to DeepSeek-R1-Zero-Qwen-32B~\citep{guo2025deepseek} in math and other reasoning tasks.  General-Zero-Qwen3-14B further improves the performance by average 7.8\% in math reasoning tasks and surpasses the other zero-RL counterparts. Furthermore, as previous zero-RL works did not explore the performance on general tasks with open-ended question answer and creative writing, we compare our General-Zero models with Qwen3-Instruct models~\citep{yang2025qwen3} on general alignment benchmarks such as Arena-Hard~\citep{li2024crowdsourced}, WritingBench~\citep{wu2025writingbench} and etc. The evaluation results show that we can achieve general abilities comparable to Qwen3-Instruct models~\citep{yang2025qwen3} via our multi-tasks zero reinforcement learning. 

Overall, our contributions in this paper can be summarized as follows: 
\begin{itemize}
    \item We propose a unified zero reinforcement learning (zero-RL) framework that integrates both verifiable and non-verifiable tasks, enabling the elicitation of reasoning capabilities across a broad range of domains, extending beyond traditional reasoning tasks.
    \item We introduce a novel length penalty tailored for general-domain data, which mitigates reward hacking problems and stabilizes the zero-RL training process.
    \item We demonstrate that reasoning behaviors acquired through our multi-task zero-RL paradigm can generalize effectively across diverse task domains.
\end{itemize}

\section{Related Works}
\subsection{Zero Reinforcement Learning in LLMs}
Large reasoning models have significantly improved the capabilities of large language models in solving complex problems. Following this direction, representative works such as DeepSeek-R1~\citep{guo2025deepseek} and OpenAI-O1~\citep{OpenAI} have demonstrated strong capabilities in mathematics and programming. In particular, DeepSeek-R1-Zero~\citep{guo2025deepseek} demonstrated that directly applying reinforcement learning on the base model can effectively uncover significant reasoning capabilities without supervised fine-tuning. This zero reinforcement learning paradigm has achieved a great success in the domain of Reinforcement Learning with Verifiable Rewards (RLVR) ~\citep{zeng2025simplerl,yu2025dapo}. Recently, several works tend to explore zero-RL in diverse general domains, such as General-Reasoner~\citep{ma2025general}, Nemotron-Crossthink~\citep{akter2025nemotron} and RLMT~\citep{bhaskar2025language}. Both General-Reasoner~\citep{ma2025general} and Nemotron-Crossthink~\citep{akter2025nemotron} incorporated multi-domain corpora in zero-RL to improve the capabilities in the general reasoning domain, where problems have well-defined ground truths and a generative verifier model is exploited to give the verifiable rewards. Although RLMT~\citep{bhaskar2025language} extended zero-RL to open-ended tasks without verifiable answers and demonstrated its potential for building better chat models, it did not address how to stabilize or improve the effectiveness of such RL process. Different from these works, we focus on enhancing reasoning abilities across math reasoning, general reasoning, and general tasks via zero reinforcement learning, and introduce strategies to stabilize the multi-task zero-RL training process.

\subsection{Generalization of Reinforcement Learning}
While most works on zero reinforcement learning have focused on mathematics or programming domains, a growing body of research has shown that reinforcement learning in verifiable domains can generalize to other fields. For instance, DeepSeek-R1~\citep{guo2025deepseek} exhibits remarkably superior performance in creative writing, even though the majority of reinforcement learning training for this model was conducted in reasoning domains. UniReasoner~\citep{huan2025does} demonstrated that zero-RL training based on mathematical reasoning can generalize to a wide range of general domains, whereas supervised fine-tuning exclusively on mathematical tasks yields limited generalization capabilities. Additionally, recent studies have further examined the cross-domain performance of reasoning models, providing deeper insights into this phenomenon~\citep{sun2025omega, hu2025breaking, cheng2025revisiting}.

\section{Methodology}
\label{headings}

\begin{figure}
    \centering
    \includegraphics[width=0.95\linewidth]{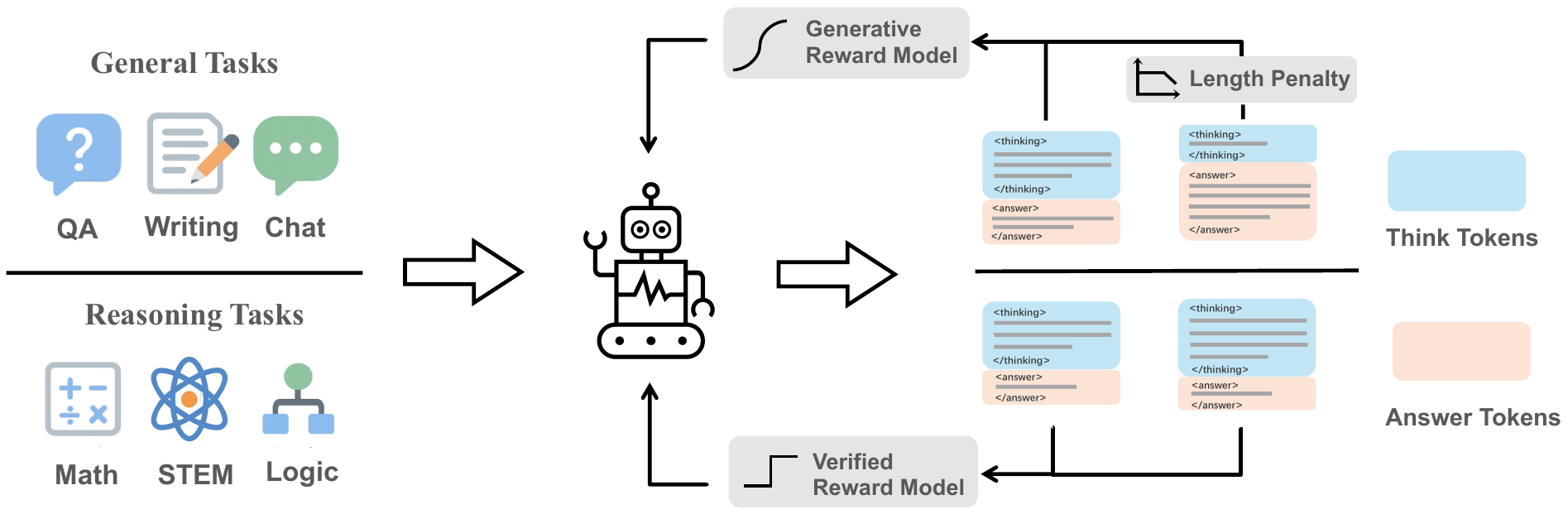}
    \caption{Overview of our unified General Zero-RL framework. The framework performs multi-task learning over both general and reasoning tasks. To mitigate reward hacking in generative reward models, a length penalty is applied when the output answer exceeds the length of the generated thinking tokens.}
    \label{fig:general_rl_framework}
\end{figure}

In this section, we provide an overview of the framework for our unified General Zero Reinforcement Learning (General Zero-RL). While prior works on zero-RL have primarily focused on verifiable tasks, this study aims to enhance models’ capabilities across general domains using zero reinforcement learning. To achieve this objective, we integrate both verifiable and non-verifiable tasks via multi-task zero reinforcement learning.

\subsection{Multi-Task Zero Reinforcement Learning}
As demonstrated by DeepSeek-R1-Zero~\citep{guo2025deepseek}, reasoning behaviors readily emerge through pure reinforcement learning on mathematical tasks, an phenomenon termed the "Aha Moment" in zero reinforcement learning. However, in our empirical study we observed that such “Aha Moment” rarely appeared when training solely on general open-ended data with a generative reward model; examples are shown in Appendix \ref{sec: appendix data examples}. One key reason is that binary rule-based rewards provide high-quality and stable learning signals on verifiable tasks, whereas model-based rewards for non-verifiable tasks are prone to the reward hacking problem, making large-scale reinforcement learning unstable or ineffective. Besides, \citet{zeng2025simplerl} further suggests that the emergence of reasoning behaviors is tightly coupled with the pre-training corpus of the base model. Since reasoning-heavy corpus mainly resides in STEM domains, eliciting comparable reasoning behaviors in open-ended tasks becomes markedly more challenging.

To alleviate these problems, we propose to integrate both general data and reasoning data in the zero-RL training process as shown in Figure~\ref{fig:general_rl_framework}, with the aim of allowing the reasoning capabilities developed during reasoning RL training to be transferred to a broader range of general domains. Specifically, we include data related to mathematics and STEM tasks with ground truth answers as verifiable tasks, where the binary verifiable reward signals are exploited. For general data, we adopt general domain prompts covering a wide range of fields including writing, question answering, casual conversation, instruction following, etc. We train a generative reward model to generate reward signals that align with human preferences. When provided with prompts and corresponding responses, the reward model outputs a scalar value to indicate the overall quality.

Formally, given blended prompts spanning both reasoning and general domains, we adopt Group Relative Policy Optimization (GRPO)~\citep{shao2024deepseekmath} as our reinforcement learning algorithm, along with two widely used modifications~\citep{yu2025dapo, drgrpo, skywork_o1}. Specifically, (1) we employ a token-level policy gradient loss rather than a sequence-level loss, and (2) we remove the KL divergence term. In each iteration, a group of responses $\{o_i\}^{G}_{i=1}$ are sampled from the policy model $\pi_{\theta_{\text{old}}}$, and the modified GRPO algorithm updates the model as follows:
\begin{equation}
\begin{split}
    \mathcal{J}_{GRPO}(\theta) &= \mathbb{E}_{q \sim P(Q), \{o_i\}_{i=1}^G \sim \pi_{\theta_{old}}(O|q)}  \\
    & \left[\frac{1}{\sum_{i=1}^{G}|o_{i}|}\sum_{i=1}^G \sum_{t=1}^{|o_i|} \min \left( r_{i,t}(\theta) A_{i,t}, \text{clip} \left( r_{i,t}(\theta), 1 - \epsilon, 1 + \epsilon \right)  A_{i,t} \right)\right] \\
    & \textrm{where}\ \ \  r_{i,t}(\theta) = \frac{\pi_{\theta}(o_{i,t} |q, o_{i, <t})}{\pi_{\theta_{old}}(o_{i,t} |q, o_{i, <t})}
\end{split}
\label{eq:GRPO-obj}
\end{equation}

The reward modeling of zero reinforcement learning usually consists of the format reward and the accuracy reward as in DeepSeek-R1-Zero~\citep{guo2025deepseek}, which will be introduced in detail in Section~\ref{sec:rewawrd modeling}. To prompt the base model to generate responses that adhere to the specified format, we employ a system prompt as shown in Table \ref{table:training templates} as the training template.

 \begin{table}
 \caption{Training template for our zero reinforcement learning, which is modified from the template of DeepSeek-R1-Zero~\citep{guo2025deepseek}.}
 \begin{center}
 \label{table:training templates}
 \begin{tabular}{l}
 \toprule
 A conversation between User and Assistant. The user asks a question, and the Assistant solves it. \\
 The assistant first  thinks about the reasoning process in the mind and then provides the user with  \\
 the answer. The reasoning process and answer are enclosed within <thinking> </thinking> and \\
 <answer> </answer> tags, respectively, i.e., <thinking> reasoning process here </thinking> \\
 <answer> answer here </answer>. Now the user asks you to solve a problem. After thinking,  \\ 
when you finally reach a conclusion, give a summary of the thinking process and clearly state \\
the conclusion within <answer> </answer> tags. \\
 \bottomrule
 \end{tabular}
 \end{center}
 \vspace{-0.1 in}
 \end{table}

\subsection{Reward Modeling}
\label{sec:rewawrd modeling}
The reward signal is vital to the optimization of reinforcement learning. We adopt both the format reward and the accuracy reward following DeepSeek-R1-Zero~\citep{guo2025deepseek}. In contrast, instead of relying solely on rule-based rewards, we decompose the accuracy reward into two distinct types, which correspond to verifiable tasks and non-verifiable tasks respectively.

\textbf{Accuracy Reward:} For verifiable tasks, the verifier-based binary reward is adopted, the accuracy reward evaluates the correctness of the responses $a$ corresponding to the ground truth answer $a_{ref}$. 
\begin{equation}
    R_{acc}^{ver}(a) = \begin{cases}
        1, & \text{if equal ($a$, $a_{ref}$)} \\
        -1, & \text{otherwise}
    \end{cases}
\end{equation}
While for open-ended tasks, a generative reward model based on Qwen2.5-72B~\citep{qwen2.5} is trained to give a scalar value ranging from -5 to 5, when provided with a question-answer pair $(q, a)$. 
\begin{equation}
    R_{acc}^{non-ver}(a) = P(q,a)
\end{equation}

\textbf{Format Reward:} The format reward ensures the response is structured according to the predefined format as Table~\ref{table:training templates}.
\begin{equation}
    R_{format}(a) = \begin{cases}
        1, & \text{if F(a)} \\
        0, & \text{otherwise}
    \end{cases}
\end{equation}
where $F(a)$ is True if the generated response $a$ is formatted correctly and False otherwise.

\subsection{Length Penalty for General Tasks}
As is widely recognized, model-based rewards on open-ended tasks are often susceptible to the reward hacking problem. In our initial experiments, we also observed this phenomenon. As illustrated in Figure~\ref{fig:no length penalty}, with the progression of training, the content within tags of <answer> </answer> in general tasks became increasingly lengthy, while the content in the <thinking> </thinking> tags shows no significant increase in length. This observation indicates that the model cannot acquire general reasoning capabilities in general tasks, which is not expected. To prevent such excessively long answers and promote deeper reasoning, we introduce two forms of length penalty, inspired by the soft over-length penalty in DAPO~\citep{yu2025dapo}.

Firstly, we impose a constraint on the length difference between the content of think (content within tags of <thinking> and </thinking>) and the content of answer (content within tags of <answer> and </answer>). When the length difference between the answer and the think content exceeds a specified value, a predefined punishment interval is defined. Within the interval, the longer the length difference, the greater the punishment it receives, which is formulated as follow: 
\begin{equation}
    R_{length}^{think}(a) = \begin{cases}
        0, & y \leq L_{max} - L_{cache} \\
        \frac{(L_{max} - L_{cache}) -|y|}{L_{cache}}, & L_{max} - L_{cache} < y \leq L_{max}\\
        -1, & L_{max} < y \\
    \end{cases}
\end{equation}
where $y = l_{answer} - l_{think}$ is the length difference between the content of answer and the content of think, $L_{max}$ and $L_{cache}$ are the predefined maximum value and the predefined punishment interval. This smoothed length penalty imposes a constraint that, when the model need to increase the length of its response, it must also extend its reasoning process in tandem. This mechanism thereby encourages the model to exhibit reasoning behaviors in general tasks.

Secondly, we also impose a length penalty on the answer content to prevent it from excessively increasing in length. We first define predefined minimum and maximum token lengths ($L_{min}^{'}$ and $L_{max}^{'}$) for the answer content. Specifically, when the length of the answer content exceeds $L_{min}^{'}$, a length penalty positively correlated with the answer’s actual length is imposed. Once the length of the answer content surpasses $L_{max}^{'}$, the length penalty is fixed at -1. This length penalty for the answer content is formally defined as follows:
\begin{equation}
  R_{length}^{answer}(a) = \begin{cases} 
    0, & l_{answer} \leq L_{min}^{'} \\
    -\lambda * l_{answer},  & L_{min}^{'} < l_{answer} \leq L_{max}^{'} \\
    -1,  & L_{max}^{'} < l_{answer}
    \end{cases}
\end{equation}
Thus, the overall length reward is defined as:
\begin{equation}
    R_{length}(a) = R_{length}^{think}(a) + R_{length}^{answer}(a)
\end{equation}

Overall, the reward signals of our General Zero-RL comprise three components: accuracy reward, format reward, and length penalty. The rewards for verifiable tasks are defined as:
\begin{equation}
R^{ver} = R_{acc}^{ver} + \alpha * R_{format}
\end{equation}
while the rewards for non-verifiable tasks are:
\begin{equation}
R^{non-ver} = R_{acc}^{non-ver} + \alpha * R_{format} + \beta * R_{length}
\end{equation}
where $\alpha$ and $\beta$ are the weighting coefficients of the format reward and the length penalty respectively.

\section{Experiments}
\label{others}

\subsection{Training Details}
\label{sec: training details}
\subsubsection{Hyper-Parameters}
We conduct experiments on Qwen3-8B-Base and Qwen3-14B-Base~\citep{yang2025qwen3} using the veRL framework\footnote{\url{https://github.com/volcengine/verl}}, an open-source reinforcement learning (RL) framework. The models are trained with a constant learning rate of 1e-6, a batch size and PPO mini-batch size of 128, and a maximum context length of 24,576 tokens. Notably, following the approach in OctoThinker~\citep{wang2025octothinker}, we gradually expand the maximum window size from 2,048 to 24,576 tokens during training. This strategy effectively prevents excessive growth in model response length during zero-RL training while accelerating the training process. In each iteration, 16 rollouts are performed per prompt with temperature and top-p both set to 1.0, and the KL coefficient was set to 0. Our General Zero-RL models are trained for approximately maximum 700 steps. For the length penalty, the predefined punishment interval $L_{max}$ and $L_{cache}$ are set to 2,048 and 1,536 tokens respectively; the predefined minimum and maximum token length for the answer content ($L_{min}^{'}$ and $L_{max}^{'}$) are set to 1,024 and 4,096 tokens respectively. The coefficient for the format reward $\alpha$ is 0.5, the length penalty coefficient $\beta$ is 2, and  $\lambda$ is set to 0.00025.

\subsubsection{Training Data}
\begin{wraptable}{r}{0.5\textwidth}
  \caption{Composition of Training Data}
 \begin{center}
 \label{table: training data}
 \small
 \begin{tabular}{c|c|c}
 \toprule
  \textbf{Data Type} & \textbf{Source} & \textbf{\#Prompts} \\
 \midrule
 Mathematics & In-house & 178,535 \\
 STEM & WebInstruct & 125,798 \\
 General & ShareGPT & 36,125 \\ 
 \bottomrule
 \end{tabular}
 \end{center}
 \end{wraptable}
In terms of training data usage, we collect approximately 178,535 in-house math-related RL prompts and we adopt the open-source WebInstruct~\citep{ma2025general} dataset as STEM-related data. Since the original WebInstruct dataset contains over 230,000 samples of different qualities, we employ the Qwen3-8B-Instruct model~\citep{yang2025qwen3} to filter out samples that are either too easy or too difficult. Specifically, we sample all data within this dataset 8 times, subsequently, we filter out samples that are either fully correct or fully incorrect, ultimately retaining 125,798 samples in our zero-RL training. For the general data portion, we use 36,125 open-source prompts from the ShareGPT dataset~\citep{sharegpt} covering a wide range of realistic user conversations. Specifically, for multi-turn conversations in ShareGPT, we split each turn into an individual sample, filtering out overly simplistic turns while retaining the preceding conversation as context. The composition of training data is summarized in Table \ref{table: training data}.

\subsection{Evaluation}
\subsubsection{Evaluation Benchmarks}
To comprehensively evaluate the model's general capabilities, we adopt the benchmarks covering \textbf{math reasoning}, \textbf{general reasoning} and \textbf{general tasks}. For math reasoning tasks, we include MATH-500~\citep{lightman2023let}, AIME24, AIME25~\citep{AIME} and OlympiadBench~\citep{he2024olympiadbench} with mathematical problems only, as standard evaluation benchmarks. For general reasoning tasks, MMLU-Pro~\citep{wang2024mmlu} is adopted as a massive multi-task benchmark to evaluate the general reasoning capability. GPQA-Diamond~\citep{rein2024gpqa} and SuperGPQA~\citep{du2025supergpqa} are two challenging benchmarks to evaluate STEM reasoning ability while BBEH~\citep{kazemi2025big} is a new benchmark extending BIG-Bench Hard~\citep{suzgun2022challenging} for better evaluation of complex reasoning. To further assess the model’s general capabilities, we utilize Arena-Hard~\citep{li2024crowdsourced}, WritingBench~\citep{wu2025writingbench}, WildBench v2~\citep{lin2024wildbench} and AlpacaEval2.0~\citep{dubois2024length} to evaluate its general alignment, creative writing, and other capabilities. During evaluation, we configure the sampling hyper-parameters as follows: temperature = 0.7, top-p = 0.8, top-k = 20, and repetition penalty = 1.05. For all the benchmarks, we set the maximum output sequence length to 24,576 tokens. For AIME24 and AIME25~\citep{AIME}, we sample 64 times for each question and take the average accuracy as the final score. For AlpacaEval2.0~\citep{dubois2024length}, we report the length-controlled win-rate as recommended with GPT-4.1 as the judge model.

\subsubsection{Comparison Baselines}

We primarily compare our models against General-Reasoner~\citep{ma2025general}, Nemotron-CrossThink~\citep{akter2025nemotron}, and UniReason~\citep{huan2025does}, all of which are trained exclusively on verifiable tasks such as mathematics or STEM-related domains. We further include DeepSeekR1-Zero-Qwen-32B~\citep{guo2025deepseek} as a strong baseline for zero-RL training on mathematical data. For general tasks, we evaluate General-Reasoner and UniReason using their publicly released models. Beyond zero-RL models, we also report results for Qwen3-Instruct~\citep{yang2025qwen3}, which is distilled from significantly larger teacher models.

% There are several works focusing on improving general reasoning capability via zero-RL training, including General-Reasoner~\citep{ma2025general}, Nemotron-CrossThink~\citep{akter2025nemotron} which are our main baselines. Besides, several other works have focused on exploring the generalization of math reasoning model to other general domains, such as UniReason~\citep{huan2025does}, we also include this work as a baseline for comparison. In the domain of zero reinforcement learning for mathematics, DeepSeekR1-Zero-Qwen-32B~\citep{guo2025deepseek} serves as a strong baseline, we thus compare our proposed models with this baseline in the reasoning domain. Furthermore, as previous works on zero reinforcement learning did not explore the performance on general tasks, we test the performance of General-Reasoner~\citep{ma2025general} and UniReason~\citep{huan2025does} on the general benchmarks with their released models. Besides, Qwen3-Instruct model~\citep{yang2025qwen3} serves as a strong baseline for general capability evaluation, as it has undergone comprehensive SFT and RL training across all domains with high-quality data.

\subsection{Main Results}

% [!thbp]
% [!htbp]
 \begin{table}
 \caption{Performance comparison of General-Zero-Qwen3-8B and General-Zero-Qwen3-14B with other counterparts across math reasoning, general reasoning, and general tasks.}
 \begin{center}
 \label{table:main results}
 \small
 \begin{tabular}{l|c c c c}
 \toprule
 \multicolumn{5}{c}{\textbf{Math Reasoning}}\\
 \midrule
  \textbf{Model} & \textbf{MATH-500} & \textbf{AIME24} & \textbf{AIME25} & \textbf{Olympiad} \\
 \midrule
 Qwen3-8B-Instruct (Non-thinking) & 87.4 &  29.1 & 20.9 & 49.6 \\
 General-Reasoner-7B & 76.0 & 13.8 & 10.4 & 37.9  \\
 General-Zero-Qwen3-8B(Ours) & \textbf{92.0} & \textbf{46.0} & \textbf{26.2} & \textbf{60.6} \\
\midrule
 Qwen3-14B-Instruct (Non-thinking) & 90.0 & 31.7 & 23.3 & 52.4 \\
 General-Reasoner-Qwen3-14B & 83.5 & 24.4 & 19.2 & 51.9  \\
 UniReason-Qwen3-14B & 87.8 & 55.7 & 38.0 & 33.8 \\
 DeepSeek-R1-Zero-Qwen-32B & 91.6 & 47.0 & -- & -- \\
 General-Zero-Qwen3-14B(Ours) & \textbf{92.4} & \textbf{59.7} & \textbf{38.2} & \textbf{73.8} \\
 \midrule
 \multicolumn{5}{c}{\textbf{General Reasoning}}\\
 \midrule
  \textbf{Model} & \textbf{MMLU-Pro} & \textbf{GPQA-D} & \textbf{SuperGPQA} & \textbf{BBEH} \\
 \midrule
 Qwen3-8B-Instruct (Non-thinking) & 66.5 & 39.3 &  36.5 & 15.3 \\
 Nemotron-CrossThink-7B & 57.8 & 38.5 & 29.1 & -- \\
 General-Reasoner-7B & 58.9  & 38.8 & 34.2 &  12.5 \\
 General-Zero-Qwen3-8B(Ours) & \textbf{68.2} & \textbf{53.0} & \textbf{39.5} & \textbf{16.3} \\
 \midrule
 Qwen3-14B-Instruct (Non-thinking) & \textbf{70.9} & 54.8 & 39.8 & 19.2 \\
 General-Reasoner-Qwen3-14B & 70.3 & 56.1 & 39.9 & 17.3  \\
 UniReason-Qwen3-14B & -- & 57.7 & -- & -- \\
 DeepSeek-R1-Zero-Qwen-32B & -- & 55.0  & -- & -- \\
 General-Zero-Qwen3-14B(Ours) & 70.6 & \textbf{58.0} & \textbf{45.3} & \textbf{20.5} \\
 \midrule
 \multicolumn{5}{c}{\textbf{General Tasks}}\\
 \midrule
  \textbf{Model} & \textbf{Arena-Hard} & \textbf{WritingBench} & \textbf{WildBenchv2} & \textbf{AlpacaEval2.0} \\
 \midrule
 Qwen3-8B-Instruct (Non-thinking) & 79.6 & 7.2 & 7.6 & 54.6 \\
 % Qwen3-8B-Instruct (Thinking) & 85.8 & 7.6 &  & 7.8 \\
 General-Reasoner-7B & 38.3 & 4.6 & 6.0 & 12.3 \\
 General-Zero-Qwen3-8B(Ours) & \textbf{86.2} & \textbf{7.7} & \textbf{7.7} & \textbf{61.9} \\
 \midrule
 Qwen3-14B-Instruct (Non-thinking) & 86.3 & 7.2 & 7.8 & 63.6 \\
 % Qwen3-14B-Instruct (Thinking) &  \textbf{91.7} & 7.8 & \textbf{8.0} &  \\
 General-Reasoner-Qwen3-14B & 76.5 & 6.3 & 7.4 & 50.6 \\
 UniReason-Qwen3-14B & 76.9 & 1.2 & 6.9 & 41.1 \\
 General-Zero-Qwen3-14B(Ours) & \textbf{89.3} & \textbf{8.5} & \textbf{8.0} & \textbf{65.0} \\
 \bottomrule
 \end{tabular}
 \end{center}
 \end{table}

 Table \ref{table:main results} summarizes the main results of our General-Zero-Qwen3-8B and General-Zero-Qwen3-14B compared to other counterparts. Most of the results of the compared baselines on reasoning tasks are the ones reported in their respective original papers. In the evaluation of reasoning tasks, our model outperforms General-Reasoner~\citep{ma2025general}, Nemotron-CrossThink~\citep{akter2025nemotron}, and UniReason~\citep{huan2025does} across math reasoning and general reasoning benchmarks. Specifically, our General-Zero-Qwen3-8B achieves 46.0\% in AIME24 surpassing General-Reasoner-7B by 32.2\% and Qwen3-8B-Instruct (non-thinking) by 16.9\%. In the evaluation of general reasoning tasks, General-Zero-Qwen3-8B achieves 68.2\% in MMLU-Pro, 53.0\% in GPQA-Diamond, 39.5\% in SuperGPQA and 16.3\% in BBEH, which consistently outperforms General-Reasoner-7B and Nemotron-CrossThink-7B significantly while also outperforming Qwen3-8B-Instruct (non-thinking) by an average of 4.8\%. In particular, the performance of our General-Zero-Qwen3-8B in reasoning tasks is comparable with DeepSeek-R1-Zero-Qwen-32B~\citep{guo2025deepseek}, a strong math reasoning baseline. For the model initialized with Qwen3-14B-Base, our General-Zero-Qwen3-14B achieves the best overall results. General-Zero-Qwen3-14B reaches 92.4\% in MATH-500, 59.7\% in AIME24 and 38.2\% in AIME25, outperforming UniReason-Qwen3-14B, a pure math reasoning model. In general reasoning tasks, General-Zero-Qwen3-14B also achieves the best performance compared to General-Reasoner-Qwen3-14B and UniReason-Qwen3-14B, both of which are initialized from the same base model. The evaluation results on reasoning benchmarks demonstrate that our models exhibit superior reasoning ability.

 \begin{figure}
% \centering
\subfigure[The growth of response length on reasoning data.
]{
\includegraphics[width=0.4\linewidth]{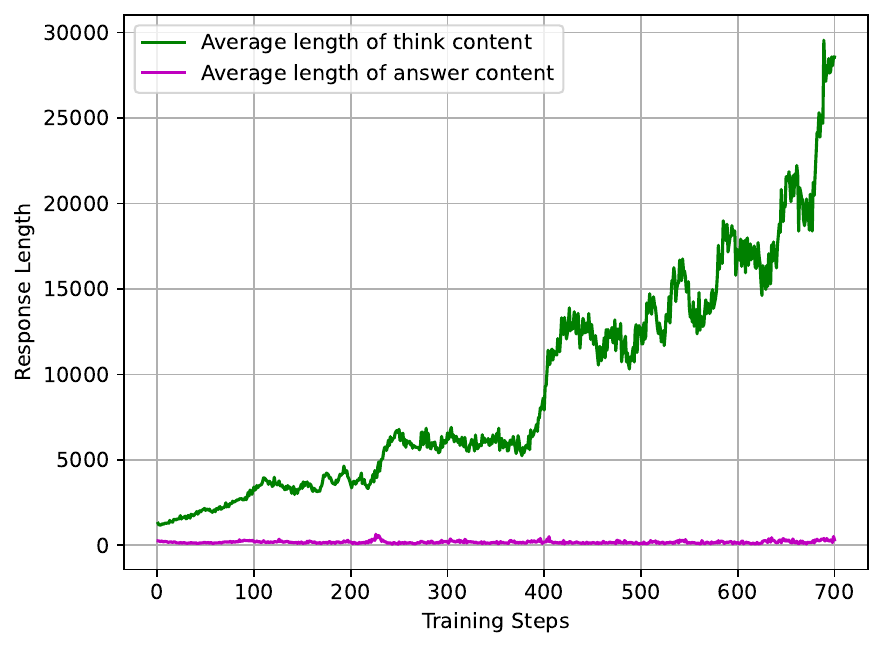}
}%
\hspace{0.05\textwidth}
\subfigure[The growth of response length on general data.]{
\includegraphics[width=0.4\linewidth]{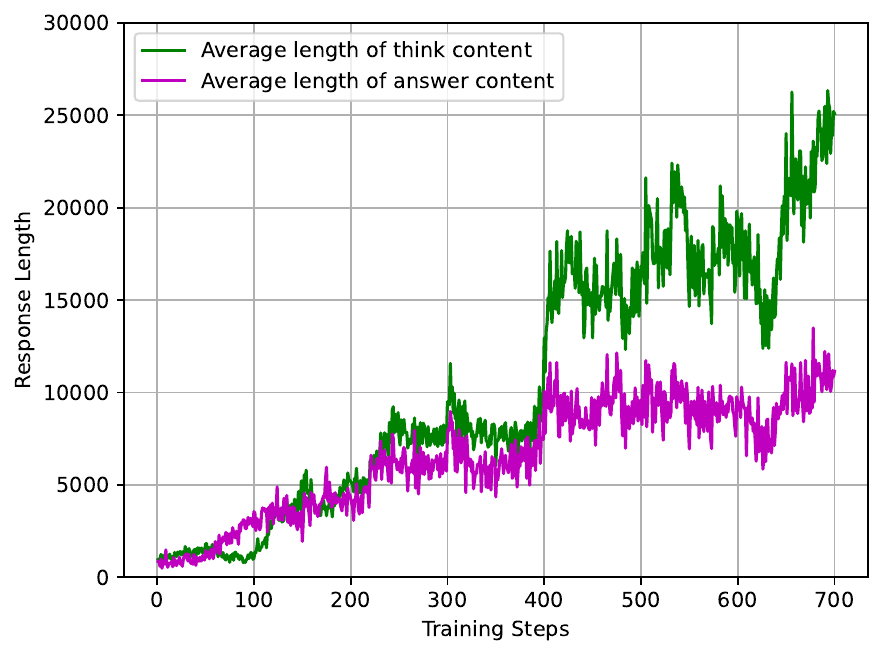}
}%
\centering
\caption{Evolution of think content length and answer length (in terms of characters) for reasoning and general tasks over the training course of General-Zero-Qwen3-14B.}
\label{fig:qwen14b_response_length}
\end{figure}

In terms of performance in general tasks, we compare our models with General-Reasoner~\citep{ma2025general}, UniReason~\citep{huan2025does} and Qwen3-Instruct (non-thinking)~\citep{yang2025qwen3}. As observed in Table \ref{table:main results}, our General Zero-RL models, across both 8B and 14B parameter sizes, outperform General-Reasoner and UniReason by a large margin on general chat and writing benchmarks, including Arena-Hard, WritingBench, WildBench v2, and AlpacaEval2.0, which demonstrates the effectiveness of our zero reinforcement learning framework. Additionally, the performance of General-Reasoner and UniReason on general tasks indicates that training exclusively on verifiable tasks fails to generalize effectively to general scenarios—a finding also corroborated by DeepSeek-R1-Zero~\citep{guo2025deepseek_nature}. Notably, our models even outperform the Qwen3-Instruct (non-thinking) models in general tasks, achieved merely through zero reinforcement learning using a small set of open-source general prompts from ShareGPT~\citep{sharegpt}. We show the growth of the length of think content and answer content on reasoning data and general data during the training process of the General-Zero-Qwen3-14B model in Figure~\ref{fig:qwen14b_response_length}. It can be observed that the length of the think content in general data continues to increase as training progresses, while the length of the answer content tends to stabilize. This phenomenon suggests that our multi-task zero-RL, together with the length penalty, effectively mitigates reward hacking and induces a more deliberate thinking process on general tasks.

 \begin{figure}
% \centering
\subfigure[Accuracy on AIME24]{
\includegraphics[width=0.4\linewidth]{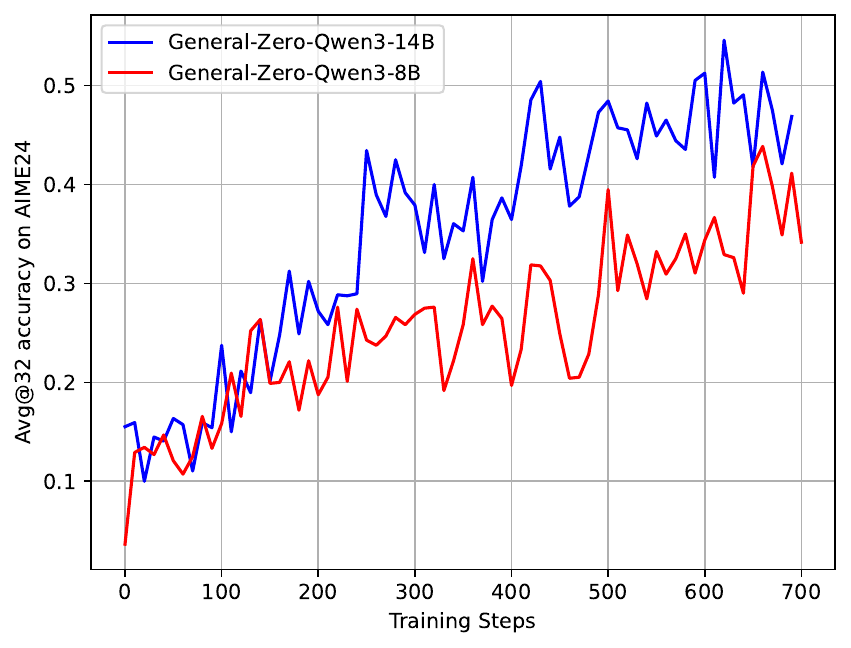}
}%
\hspace{0.05\textwidth}
\subfigure[Mean of response token length]{
\includegraphics[width=0.4\linewidth]{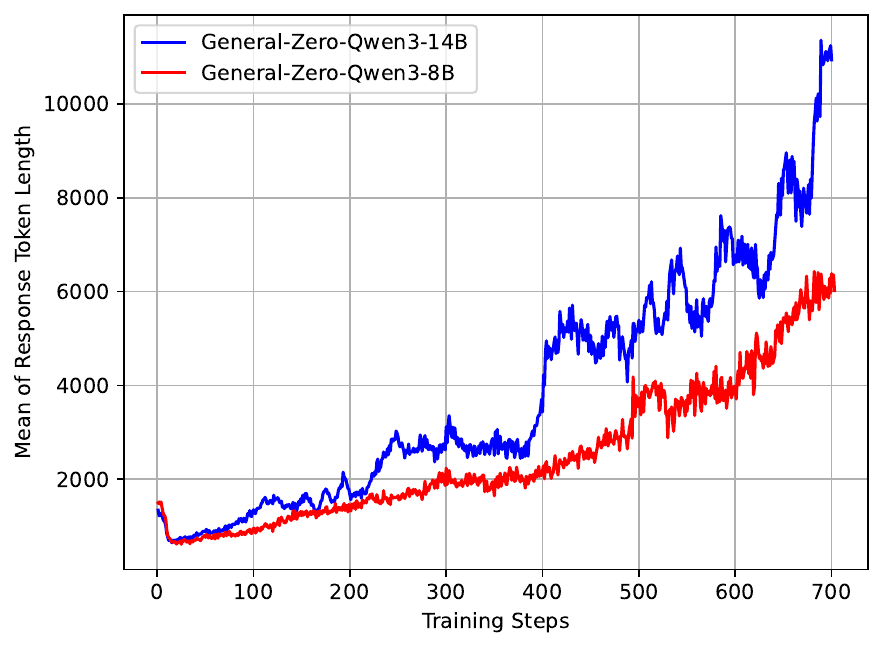}
}%
\centering
\caption{Accuracy of AIME24 and response length of General-Zero-Qwen3-14B and General-Zero-Qwen3-8B during the training process of zero reinforcement learning.}
\label{fig: qwen14b accuracy and response}
\end{figure}

In Figure \ref{fig: qwen14b accuracy and response}, we illustrate the training dynamics of General-Zero-Qwen3-8B and General-Zero-Qwen3-14B models, where the growth of response length and the evaluation metrics on AIME24 during the training processes are displayed. We can observe that as training progresses, the model's accuracy on AIME24 consistently improves, while the average response length on the training set also increases steadily. Additionally, the General-Zero-Qwen3-14B model generates significantly longer responses than the General-Zero-Qwen3-8B model with consistently better performance, which indicates that a more powerful base model is correlated with stronger reasoning capabilities.

\subsection{Ablation Study}

\subsubsection{Ablation of Multi-Task Training}

\begin{table}[!htbp]
  \caption{Ablation study of reasoning-only training and multi-task training. }
 \begin{center}
 \label{table:ablate math results}
 \small
 \begin{tabular}{l|c| c |c | c }
 \toprule
  \textbf{Model} & \textbf{MATH500} & \textbf{MMLU-Pro} & \textbf{GPQA-D} & \textbf{Arena-Hard}  \\
 \midrule
 Reasoning-Only Training & 86.8 & 67.4 & \textbf{48.9} & 71.5 \\
 Multi-Task Training  & \textbf{87.8} & \textbf{67.5} & 48.0 & \textbf{84.8} \\
 \bottomrule
 \end{tabular}
 \end{center}
 \end{table}
 
To investigate improvements in general performance, we conduct an ablation study comparing reasoning-only training and multi-task training. All models are initialized with Qwen3-8B-Base~\citep{yang2025qwen3} and trained using zero reinforcement learning on two distinct datasets: one consisting of reasoning-only data, and the other of multi-task data (including general tasks). For reasoning-only training, only verifiable tasks are included while multi-task training incorporates verifiable and non-verifiable tasks concurrently. Training parameters (e.g., learning rate and batch size) are kept consistent with those specified in Section~\ref{sec: training details}. The models are trained for 450 steps and subsequently evaluated on several standard benchmarks. 

The results, summarized in Table~\ref{table:ablate math results}, show that both reasoning-only and multi-task trained models achieve comparable performance on reasoning benchmarks such as MATH-500, MMLU-Pro, and GPQA-Diamond. However, on general-domain benchmarks (e.g., Arena-Hard), the reasoning-only trained model underperforms its multi-task counterpart by 13.3\%, highlighting that training solely on verifiable tasks does not effectively transfer reasoning ability to broader domains and underscoring the necessity of including general tasks in the RL training process. Moreover, in our empirical study, we find that training solely on general data not only fails to effectively elicit reasoning behaviors but also suffers from the severe reward hacking problem. The training dynamics comparing general data-only training and multi-task training are presented in Appendix \ref{sec: general data-only}.

\subsubsection{Ablation of Length Penalty}

 \begin{figure}
% \centering
\subfigure[The growth of response length with length penalty.
]{
\includegraphics[width=0.4\linewidth]{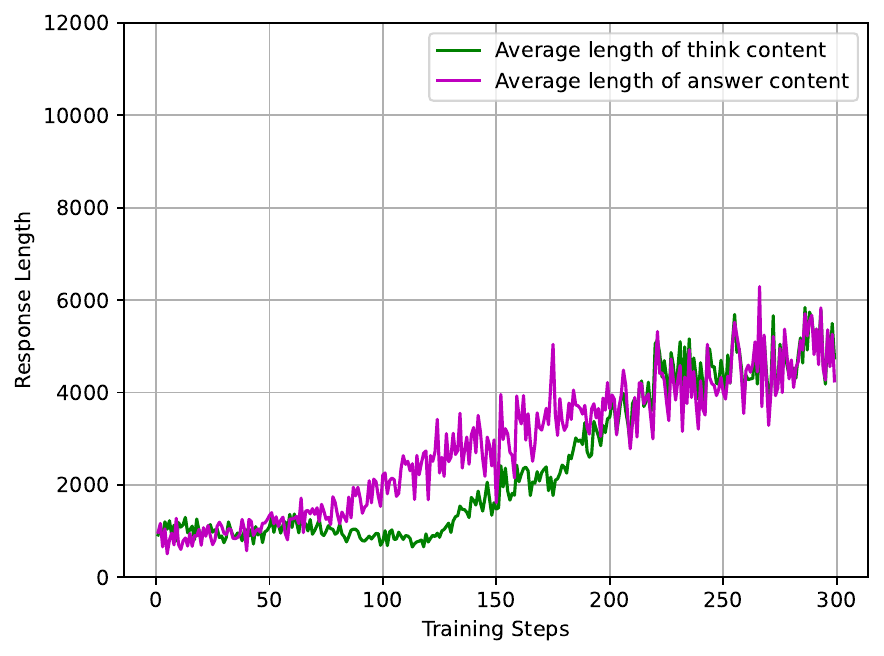}
}%
\hspace{0.05\textwidth}
\subfigure[The growth of response length without length penalty.]{
\label{fig:no length penalty}
\includegraphics[width=0.4\linewidth]{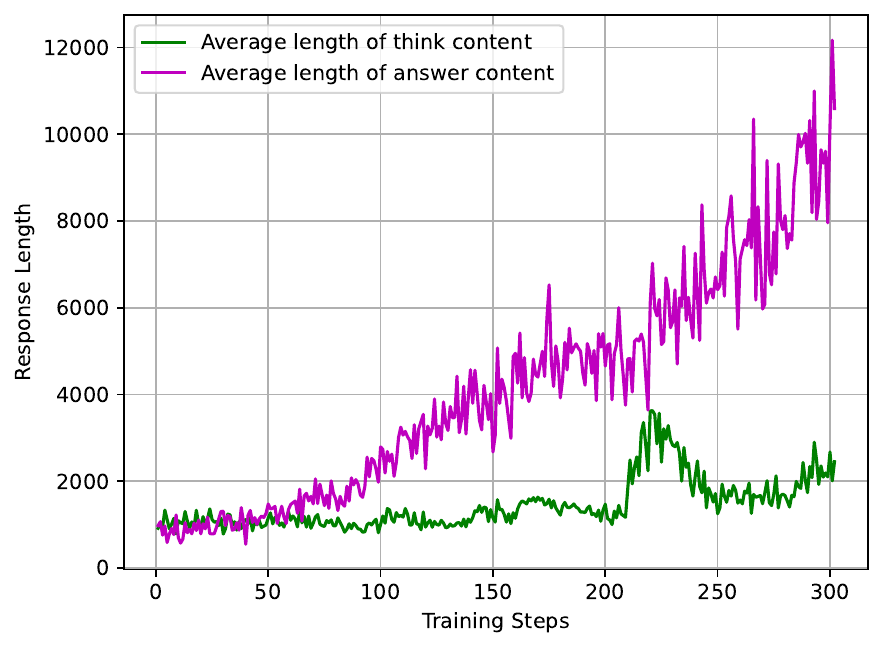}
}%
\centering
\caption{Evolution of think content and answer lengths (in terms of characters) on general data throughout the training of General-Zero-Qwen3-8B, comparing with and without length penalty.}
\label{fig:length penalty}
\end{figure}

Another critical component of our algorithm is the design of the length penalty. To validate the effectiveness of our proposed length penalty on general-domain data, we compare the growth trends of the think content length and the answer content length (in terms of characters) during the zero-RL training process in Figure~\ref{fig:length penalty}. As observed in subfigure (b), without the length penalty, the average length of the think content remains largely unchanged while the length of the answer content grows rapidly, suggesting that the model’s reasoning behavior is not effectively elicited and the reward hacking problem occurs to favor longer responses. In contrast, subfigure (a) shows that when the length penalty is applied, the lengths of both the think content and answer content exhibit coordinated and reasonable growth. Additionally, the average length of answer content is shorter than in the scenario without the length penalty (comparing the purple curve in subfigure (a) and (b)), underscoring the necessity of incorporating the length penalty into general zero reinforcement learning.

% we show in Figure~\ref{fig:length penalty} the growth trends of think-trace and answer lengths on general data during the training of General-Zero-Qwen3-8B, comparing scenarios with and without the length penalty. Without the length penalty, the average think-trace length remains largely unchanged while the answer length grows rapidly, suggesting that the model’s reasoning behavior is not effectively stimulated and that reward hacking occurs. In contrast, with the length penalty, both think-trace and answer lengths increase in tandem, highlighting the necessity of incorporating length penalty in general zero reinforcement learning.

% \subsubsection{Effect of the Maximum Window Size}
% \label{sec: effect of the Maximum Window Size}

\subsection{Limitations}
In this paper, we investigate zero reinforcement learning in broader domains, including math reasoning, general reasoning and general tasks. However, we did not show the results on benchmarks related to programming, as we did not include the code-related tasks and reward models in our training. Since programming is a relatively specialized domain, its reward signals require components like code sandboxes, and such elements would increase the complexity of our algorithm. In this paper, we mainly focus on how to incorporate general data into zero-RL training process, so we do not include code-related training. However, numerous works (e.g., Absolute zero~\citep{zhao2025absolute}, Coder-R1~\citep{liu2025code}, AceCoder~\citep{zeng2025acecoder}) have demonstrated that reasoning behaviors in programming domain can be acquired through zero reinforcement learning. Therefore, exploring how to integrate code-related data in multi-task zero-RL training can be pursued as future work.

Additionally, we do not compare our model’s performance with Qwen3-Instruct models in thinking mode. This is because Qwen3-Instruct’s thinking mode is typically trained on chains-of-thought (CoT) generated by much larger models during the supervised fine-tuning (SFT) phase. In contrast, the chain-of-thought capabilities of our zero-RL trained model based on Qwen3-14B-Base are not yet sufficient to match the CoT produced by larger models, as also observed in DeepSeek-R1~\citep{guo2025deepseek}. In future work, we will extend our zero-RL training to larger base models and compare them with more competitive reasoning models.

\subsection{Conclusion}
In this paper, we propose a multi-task zero reinforcement learning algorithm to incorporate general data into zero-RL training. By applying a smooth length penalty on general data, we effectively mitigate the reward hacking problem. In comprehensive evaluations, our models achieve superior reasoning and general performance compared to other zero-RL baselines, while also demonstrating comparable general performance with Qwen3-Instruct models. Our results not only provide an effective zero-RL paradigm for stimulating reasoning ability on general tasks, but also highlight the limitations of reasoning-domain-only RL methods, underscoring the importance of including more general RL tasks in future work.


\bibliography{iclr2025_conference}
\bibliographystyle{iclr2025_conference}

\appendix
\section{Appendix}

\subsection{General Data-Only Training}
\label{sec: general data-only}

 \begin{figure}[!htbp]
% \centering
\subfigure[Average length of think content.
]{
\includegraphics[width=0.4\linewidth]{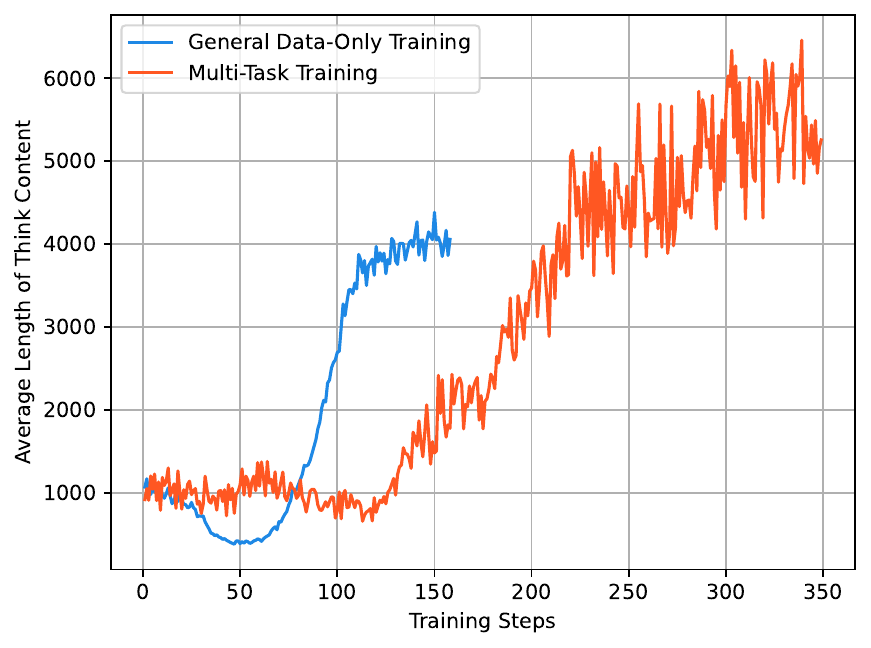}
}%
\hspace{0.05\textwidth}
\subfigure[Average length of answer content]{
\includegraphics[width=0.4\linewidth]{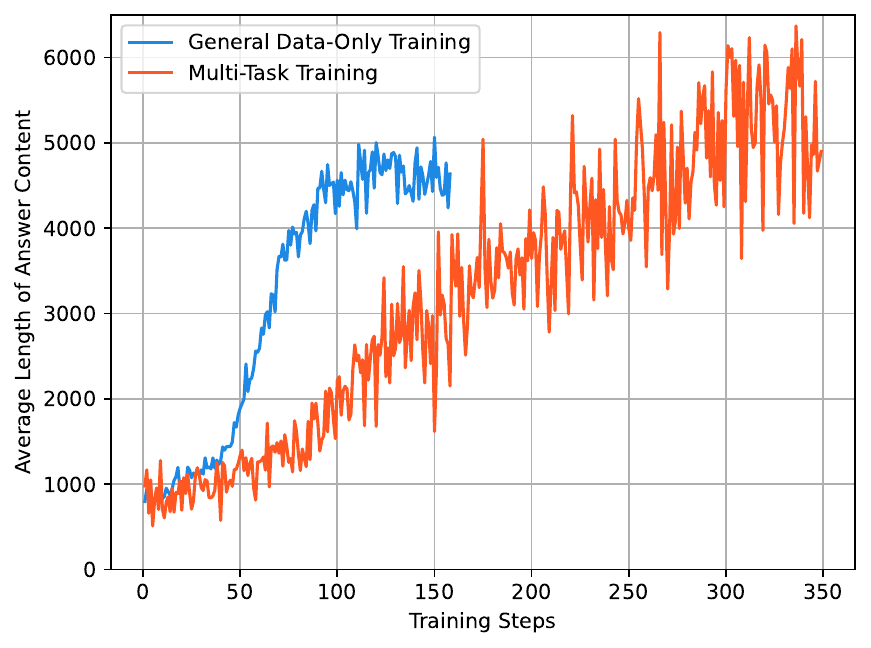}
}%
\centering
\caption{Evolution of think content length and answer length (in terms of characters) on general data during the training process of General-Zero-Qwen3-8B models when trained with general-only data and multi-task data.}
\label{fig:general data-only}
\end{figure}

In Figure \ref{fig:general data-only}, we compare the growth of think content length and answer content length  (in terms of characters) when training Qwen3-8B-Base on general-only data versus multi-task data with length penalty applied. For the general-only data training, as we observed abnormal growth of think and answer lengths alongside a noticeable reward hacking problem, the training was stopped at around 150 steps. It is evident that when the model is trained exclusively on general data, response lengths increase significantly faster than in multi-task training. Furthermore, general-only data training (the blue curves in both subfigure (a) and (b)) show a sharp and abnormal increase in both think and answer content lengths after around 50 training steps, which is a typical manifestation of reward hacking—where the model inflates content length without substantive reasoning to maximize rewards. In contrast, multi-task training exhibits a continuous and gradual growth in both the think and answer content lengths, suggesting that multi-task training can mitigate the reward hacking issue to a certain degree and promote a more reasonable and coordinated length evolution that aligns with the actual reasoning needs.

\subsection{Data Examples}
\label{sec: appendix data examples}
We show several training data examples generated by models trained with general-only data and multi-task data. Case~1 is generated by a general data-only trained model while Case~2 and Case~3 are generated by the multi-task trained model (General-Zero-Qwen3-8B) respectively. Due to space constraints, only the core structure of the chain-of-thought is presented for Case~2 and Case~3, with certain contents omitted. Case~1 and Case~2 are training cases generated during the zero-RL process by different models using the same prompt. It can be observed that the chain-of-thought in Case~1 lacks a meaningful reasoning process, whereas Case~2 demonstrates a step-by-step thinking process, and is also formatted in markdown, which reveals the effectiveness of our multi-task zero reinforcement learning for eliciting reasoning behavior for general tasks. 

Both Case~2 and Case~3 are examples generated by General-Zero-Qwen3-8B in question-answering and writing scenario respectively. It can be observed that in the question answer scenario, the model provides a step-by-step analysis and ultimately delivers a comprehensive answer that covers multiple dimensions. In the creative writing scenario, however, the model first identifies the core components of email writing, offers suggestions for crafting the email from different perspectives, and finally generates a response in the form of an email as required by the question. 

% \begin{tcolorbox}[casebox,
% title={Case 1: Generated by general data-only trained model}
% ]
% \textbf{Question}: Explain the human perspective hierarchy in the context of CEOs at the top of the perspective hierarchy and individual contributors being at the bottom using clear and precise expressions of concepts and ideas. \\

% \textbf{Response}: 
% \markdownInput{example_bad.md}  % 导入同目录下的 example.md
% \end{tcolorbox}

\begin{tcolorbox}[casebox,
title={Case 1: Generated by general data-only trained model}
]
\textbf{Question}: Explain the human perspective hierarchy in the context of CEOs at the top of the perspective hierarchy and individual contributors being at the bottom using clear and precise expressions of concepts and ideas. \\

\textbf{Response}: 
\textbf{<thinking>}
The task is to explain the human perspective hierarchy, focusing on the roles of CEOs and individual contributors. This involves understanding and articulating the differences in responsibility, decision-making influence, and organizational impact between these two levels. The explanation should be clear and precise, avoiding ambiguity.
\textbf{</thinking>}

\textbf{<answer>}
The human perspective hierarchy, within an organizational context, delineates the different roles and responsibilities individuals occupy based on their influence and decision-making power. At the top of this hierarchy are CEOs (Chief Executive Officers), who hold the highest level of authority and responsibility. They are responsible for setting the overall vision, strategy, and goals of the organization, making crucial decisions that impact the company's direction and success. CEOs are at the pinnacle because they have significant influence over major stakeholders, including shareholders, boards, and senior management teams.

Moving down the hierarchy, we encounter senior executives and managers who implement the strategies set by the CEO, overseeing specific departments or functions within the organization. These individuals are critical in translating broad strategic goals into actionable plans and policies.

At the middle levels are middle managers, who bridge the gap between senior leadership and individual contributors. They are responsible for managing teams, ensuring operational efficiency, and facilitating communication between different parts of the organization. Middle managers play a pivotal role in aligning team efforts with organizational objectives.

Finally, at the bottom of the perspective hierarchy are the individual contributors—employees who perform specific tasks and roles. They are the backbone of the organization, directly involved in the day-to-day operations that bring the business to life. Despite being at the lowest rung, their contributions are essential and often undervalued, as they are the ones who execute the policies and processes established by higher-ups.

In summary, the human perspective hierarchy is structured such that CEOs and senior executives have overarching influence and decision-making power, while middle managers ensure alignment and operational efficiency, and individual contributors are crucial for the actual execution of work.
\textbf{</answer>}
\end{tcolorbox}

% \begin{tcolorbox}[floatbox={Case 1}{box:qa}]
% \textbf{Question}: Explain the human perspective hierarchy in the context of CEOs at the top of the perspective hierarchy and individual contributors being at the bottom using clear and precise expressions of concepts and ideas. \\

% \textbf{Response}: 
% \markdownInput{example_bad.md}  % 导入同目录下的 example.md

% \end{tcolorbox}

% 带标题的案例框
\begin{tcolorbox}[casebox, title={Case 2: Generated by multi-task data trained model}]
\textbf{Question}: Explain the human perspective hierarchy in the context of CEOs at the top of the perspective hierarchy and individual contributors being at the bottom using clear and precise expressions of concepts and ideas. \\

\textbf{Response}: 
% \markdownInput{example.md}  % 导入同目录下的 example.md
\textbf{<thinking>}
Alright, let’s break this down systematically. The goal is to explain the \textbf{human perspective hierarchy} in the context of an organization, with CEOs at the top and individual contributors at the bottom. To do this effectively, I’ll need to:

1. \textbf{Define what a "perspective hierarchy" means in this context.} \\
2. \textbf{Map how perspectives differ across levels (CEOs vs. individual contributors).} \\
3. \textbf{Use clear, precise language to articulate the differences and relationships between these perspectives.}

\rule{\textwidth}{0.6pt} 

Step 1: Understand What a \textbf{Perspective Hierarchy} Means 

...

Step 2: Define the \textbf{Human Perspective Hierarchy} in an Org Structure

...

Step 3: How Perspectives Differ Across the Hierarchy

...

Step 4: The Human Perspective Hierarchy in Action

...

Step 5: Why This Matters (and Potential Challenges)

...

Step 6: Summarize in Clear, Precise Language

...

By understanding this hierarchy, leaders can better communicate across levels, address alignment challenges, and create a more cohesive organizational culture.
\textbf{</thinking>}

\textbf{<answer>}
The Human Perspective Hierarchy: From CEOs to Individual Contributors

The \textbf{human perspective hierarchy} in an organization describes how perspectives—how people understand, interpret, and approach problems—differ across levels of the hierarchy, from CEOs at the top to individual contributors at the bottom. These differences are shaped by \textbf{role, information access, decision-making scope, and time horizon}. Here’s a clear breakdown of the perspective hierarchy in this context:

\rule{\textwidth}{0.6pt} 

\textbf{1. CEOs (Top of the Hierarchy): The Strategic Visionaries}

...

\textbf{2. Senior Executives (e.g., COOs, CFOs, CTOs): The Tactical Implementers}

...

\textbf{3. Middle Managers (e.g., Department Heads, Team Leads): The Operational Orchestrators}

...

\textbf{4. Individual Contributors (ICs): The Task Executors}

...

\rule{\textwidth}{0.6pt} 

\textbf{Key Dynamics of the Perspective Hierarchy}

...

\textbf{Challenges in the Perspective Hierarchy}

...

\textbf{How to Navigate the Perspective Hierarchy Effectively}

...

\textbf{Final Summary: The Human Perspective Hierarchy in a Nutshell}

...

\rule{\textwidth}{0.6pt} 

In essence, the \textbf{human perspective hierarchy} is a natural structure that reflects the \textbf{diverse roles and responsibilities} within an organization. While it ensures clarity and focus at each level, it also requires \textbf{deliberate effort} to maintain alignment, communication, and empathy across the hierarchy. By understanding and navigating these differing perspectives, organizations can turn complexity into coherence and ensure everyone is working toward the same goals.
\textbf{</answer>}

\end{tcolorbox}

\begin{tcolorbox}[casebox, title={Case 3: Generated by multi-task data trained model}]
\textbf{Question}: Write an email to apply a headcount backfill from leadership. Mekhi on Data Platform will leave the company in early June, and we are thinking to backfill the headcount with Zicheng, an intern who worked on Data Platform last summer but we weren't able to extend offer to due to lack of headcount. Zicheng will come in as L3 new grad, but he has good context on the team and he has high potential that can contribute to the team. \\

\textbf{Response}: 
% \markdownInput{example_a.md}  % 导入同目录下的 example.md
\textbf{<thinking>}
Alright, I need to draft an email to apply for a headcount backfill. Essentially, Mekhi, who’s leaving the Data Platform team in early June, had an intern named Zicheng who worked with the team last summer. We missed out on extending an offer to Zicheng last time due to lack of headcount, but now that we’re letting Mekhi go, we want to bring Zicheng on as a new hire (L3 new grad). He’s a great fit because he already has context on the team and shows high potential to contribute. My goal is to write a professional, persuasive email to leadership explaining why we should approve this backfill.

\textbf{Key Points to Cover in the Email:}

% 1. \textbf{Why the Backfill Makes Sense:} ...

% 2. \textbf{Zicheng’s Qualifications:} ...

% 3. \textbf{Why Now is the Right Time:} ...

% 4. \textbf{Team Impact:} ...

\begin{enumerate}
    \item \textbf{Why the Backfill Makes Sense:} ...
    \item \textbf{Zicheng’s Qualifications:} ...
    \item \textbf{Why Now is the Right Time:} ...
    \item \textbf{Team Impact:} ...
\end{enumerate}

\textbf{Drafting the Email:}

Let’s structure the email professionally. I’ll address it to the relevant leadership (maybe the Head of Engineering, Talent Partner, or someone like that—depending on the org structure). The tone should be confident but respectful, acknowledging the constraints we faced before and framing this as a no-brainer now that Mekhi is leaving.

\rule{\textwidth}{0.6pt} 

\textbf{Subject:} Request to Approve Headcount Backfill for Zicheng (Data Platform, L3 New Grad) – Opportunity to Fill Mekhi’s Departure  

...

\rule{\textwidth}{0.6pt} 

\textbf{Key Considerations While Writing:}

...

\textbf{Alternative Angle (if needed):}

...

\rule{\textwidth}{0.6pt} 

This draft feels solid, but I can adjust it further if needed. For example, if the headcount holder is someone specific (e.g., the Head of Data Platform), I might add a sentence acknowledging their role or seeking their input. Otherwise, this should cover the request effectively.
\textbf{</thinking>}

\textbf{<answer>}
Here’s a polished and professional email draft to apply for the headcount backfill for Zicheng:

\rule{\textwidth}{0.6pt} 

\textbf{Subject:} Request to Approve Headcount Backfill for Zicheng (Data Platform, L3 New Grad) – Perfect Fit for Mekhi’s Opening  

...

Best regards,  

[Your Full Name] 

[Your Job Title] 

[Your Contact Information]

\rule{\textwidth}{0.6pt} 

\textbf{Why This Works:}

...

Let me know if you’d like to tweak this further!  \textbf{</answer>}

\end{tcolorbox}

\end{document}